\pdfoutput=1
\documentclass[10pt,twocolumn,letterpaper]{article}

\usepackage{iccv}
\usepackage{times}
\usepackage{epsfig}
\usepackage{graphicx}
\usepackage{amsmath}
\usepackage{amssymb}

\usepackage{subcaption}
\usepackage{tabularx}
\usepackage{booktabs}
\usepackage[table]{xcolor}
\usepackage[nomessages]{fp} % for fixed-point arithmetics
\usepackage{capt-of}
\usepackage{etoolbox}
\usepackage[super]{nth}

\usepackage[breaklinks=true,bookmarks=false]{hyperref}
\usepackage[capitalize]{cleveref}

\iccvfinalcopy % *** Uncomment this line for the final submission

\def\iccvPaperID{****} % *** Enter the ICCV Paper ID here

\makeatletter
\renewcommand*\subcaption@label{%
    \caption@withoptargs\subcaption@@label}
\makeatother

\newcommand{\maxnum}{100.00}
\newcommand{\maxtext}{100.00}
\newcommand{\baralign}{r}
\newlength{\maxlen}
\newcommand{\databar}[3][cyan!25]{%
    \settowidth{\maxlen}{\maxtext}%
    \addtolength{\maxlen}{\tabcolsep}%
    \FPeval\result{round(#3/\maxnum:4)}%
    \rlap{\color{#1}\hspace*{-.5\tabcolsep}\rule[-.05\ht\strutbox]{\result\maxlen}{.95\ht\strutbox}}%
    \makebox[\dimexpr\maxlen-\tabcolsep][\baralign]{#2}%
}

\newcommand{\teaserfigure}{
    \begin{center}
        \begin{tabular}{ @{}
                m{0.032\linewidth} @{}
                | @{\hspace{0.014\linewidth}}
                >{\centering}m{0.09\linewidth} @{\hspace{0.014\linewidth}}
                >{\centering}m{0.09\linewidth} @{\hspace{0.014\linewidth}}
                >{\centering}m{0.09\linewidth} @{\hspace{0.014\linewidth}}
                | @{\hspace{0.014\linewidth}}
                >{\centering}m{0.09\linewidth} @{\hspace{0.014\linewidth}}
                >{\centering}m{0.09\linewidth} @{\hspace{0.014\linewidth}}
                >{\centering}m{0.09\linewidth} @{\hspace{0.014\linewidth}}
                | @{\hspace{0.014\linewidth}}
                >{\centering}m{0.09\linewidth} @{\hspace{0.014\linewidth}}
                >{\centering}m{0.09\linewidth} @{\hspace{0.014\linewidth}} >{\centering\arraybackslash}m{0.09\linewidth} @{} }
            & \multicolumn{3}{c@{\hspace{0.014\linewidth}}|@{\hspace{0.014\linewidth}}}{\large Exact Duplicate}
            & \multicolumn{3}{c@{\hspace{0.014\linewidth}}|@{\hspace{0.014\linewidth}}}{\large Near-Duplicate}
            & \multicolumn{3}{c}{\large Very Similar} \\[1ex]
            \rotatebox{90}{Test} &
            \includegraphics[width=\linewidth]{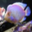} &
            \includegraphics[width=\linewidth]{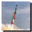} &
            \includegraphics[width=\linewidth]{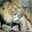} &
            \includegraphics[width=\linewidth]{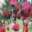} &
            \includegraphics[width=\linewidth]{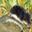} &
            \includegraphics[width=\linewidth]{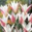} &
            \includegraphics[width=\linewidth]{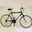} &
            \includegraphics[width=\linewidth]{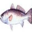} &
            \includegraphics[width=\linewidth]{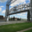} \\[1.5ex]
            \rotatebox{90}{Training} &
            \includegraphics[width=\linewidth]{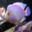} &
            \includegraphics[width=\linewidth]{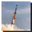} &
            \includegraphics[width=\linewidth]{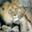} &
            \includegraphics[width=\linewidth]{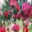} &
            \includegraphics[width=\linewidth]{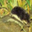} &
            \includegraphics[width=\linewidth]{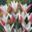} &
            \includegraphics[width=\linewidth]{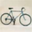} &
            \includegraphics[width=\linewidth]{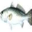} &
            \includegraphics[width=\linewidth]{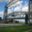} \\
        \end{tabular}%
    \end{center}%
    \vspace{-\baselineskip}\captionof{figure}{Examples for different types of duplicates between the CIFAR-100 test and training set. The top row shows images from the test set and the bottom row shows their nearest neighbors from the training set in a CNN feature space. Please see the main text for a description of the three categories of duplicates.}%
    \label{fig:dup-examples}%
}

\makeatletter
\apptocmd\@maketitle{{\vspace{-1.7\baselineskip}\teaserfigure{}\par\vspace{1.2\baselineskip}}}{}{}
\makeatother

\begin{document}

%%%%%%%%% TITLE
\title{Do we train on test data?\\Purging CIFAR of near-duplicates}

\author{Bj{\"o}rn Barz \hspace{3em} Joachim Denzler\\\\
Computer Vision Group\\
Friedrich Schiller University Jena\\
Ernst-Abbe-Platz 2, 07743 Jena, Germany\\
{\tt\small bjoern.barz@uni-jena.de}
%\and
}

\maketitle
%\thispagestyle{empty}

%%%%%%%%% ABSTRACT
\begin{abstract}
    The CIFAR-10 and CIFAR-100 datasets are two of the most heavily benchmarked datasets in computer vision and are often used to evaluate novel methods and model architectures in the field of deep learning.
    However, we find that 3.3\% and 10\% of the images from the test sets of these datasets have duplicates in the training set.
    These duplicates are easily recognizable by memorization and may, hence, bias the comparison of image recognition techniques regarding their generalization capability.
    To eliminate this bias, we provide the ``fair CIFAR'' (ciFAIR) dataset, where we replaced all duplicates in the test sets with new images sampled from the same domain.
    %The training set remains unchanged, in order not to invalidate pre-trained models.
%
    We then re-evaluate the classification performance of various popular state-of-the-art CNN architectures on these new test sets to investigate whether recent research has overfitted to memorizing data instead of learning abstract concepts.
    We find a significant drop in classification accuracy of between 9\% and 14\% relative to the original performance on the duplicate-free test set.
   The ciFAIR dataset and pre-trained models are available at \url{https://cvjena.github.io/cifair/}, where we also maintain a leaderboard.
\end{abstract}
\vspace{-2\baselineskip}

%%%%%%%%% BODY TEXT
\section{Introduction}

Almost ten years after the first instantiation of the ImageNet Large Scale Visual Recognition Challenge (ILSVRC) \cite{russakovsky2015ilsvrc}, image classification is still a very active field of research.
The majority of recent approaches belongs to the domain of \textit{deep learning} with several new architectures of \textit{convolutional neural networks} (CNNs) being proposed for this task every year and trying to improve the accuracy on held-out test data by a few percent points \cite{he2016resnet,zagoruyko2016wrn,xie2017resnext,huang2017densenet,han2017pyramidnet,real2018amoebanet,Barz18:GoodTraining}.
A key to the success of these methods is the availability of large amounts of training data \cite{krizhevsky2012alexnet,sun2017revisiting}.
The world wide web has become a very affordable resource for harvesting such large datasets in an automated or semi-automated manner \cite{deng2009imagenet,krizhevsky2009cifar,huiskes08mirflickr,tencent-ml-images-2019}.

A problem of this approach is that there is no effective automatic method for filtering out near-duplicates among the collected images.
When the dataset is split up later into a training, a test, and maybe even a validation set, this might result in the presence of near-duplicates of test images in the training set.
Usually, the post-processing with regard to duplicates is limited to removing images that have exact pixel-level duplicates \cite{krizhevsky2009cifar,deng2009imagenet}.
However, many duplicates are less obvious and might vary with respect to contrast, translation, stretching, color shift etc.
These are variations that can easily be accounted for by data augmentation, so that these variants will actually become part of the augmented training set.

For a proper scientific evaluation and model selection, the presence of such duplicates is a critical issue:
Ideally, researchers and machine learning practitioners aim at comparing models with respect to their ability of generalizing to unseen data.
With a growing number of duplicates, however, we run the risk of comparing them in terms of their capability of memorizing the training data, which increases with model capacity.
This is especially problematic when the difference between the error rates of different models is as small as it is nowadays, \ie, sometimes just one or two percent points.
The significance of these performance differences hence depends on the overlap between test and training data.
In some fields, such as fine-grained recognition, this overlap has already been quantified for some popular datasets, \eg, for the Caltech-UCSD Birds dataset \cite{WahCUB_200_2011,jaderberg2015spatial}.

In this work, we assess the number of test images that have near-duplicates in the training set of two of the most heavily benchmarked datasets in computer vision: CIFAR-10 and CIFAR-100 \cite{krizhevsky2009cifar}.
We will first briefly introduce these datasets in \cref{sec:cifar} and describe our duplicate search approach in \cref{sec:duplicate-search}.
In a nutshell, we search for nearest neighbor pairs between test and training set in a CNN feature space and inspect the results manually, assigning each detected pair into one of four duplicate categories.
We find that 3.3\% of CIFAR-10 test images and a startling number of 10\% of CIFAR-100 test images have near-duplicates in their respective training sets.

Subsequently, we replace all these duplicates with new images from the Tiny Images dataset \cite{torralba2008tinyimages}, which was the original source for the CIFAR images (see \cref{sec:new-test-set}).
To determine whether recent research results have already been affected by these duplicates, we finally re-evaluate the performance of several state-of-the-art CNN architectures on these new test sets in \cref{sec:reevaluation}.

Similar to our work, Recht et al.~\cite{recht2018cifar} have recently sampled a completely new test set for CIFAR-10 from Tiny Images to assess how well existing models generalize to truly unseen data.
Furthermore, they note parenthetically that the CIFAR-10 test set comprises 8\% duplicates with the training set, which is more than twice as much as we have found.
As opposed to their work, however, we also analyze CIFAR-100 and only replace the duplicates with new images, while leaving the remaining ones untouched.
Moreover, we employ a fine-grained distinction between three different types of duplicates.
We make the list of duplicates found, the new test sets, and pre-trained models publicly available at \url{https://cvjena.github.io/cifair/}.

\section{The CIFAR Datasets}
\label{sec:cifar}

There are two different CIFAR datasets \cite{krizhevsky2009cifar}: CIFAR-10, which comprises 10 classes, and CIFAR-100, which comprises 100 classes.
Both contain 50,000 training and 10,000 test images.
Neither the classes nor the data of these two datasets overlap, but both have been sampled from the same source: the Tiny Images dataset \cite{torralba2008tinyimages}.

In this context, the word ``tiny'' refers to the resolution of the images, not to their number.
Quite the contrary, Tiny Images comprises approximately 80 million images collected automatically from the web by querying image search engines for approximately 75,000 synsets of the WordNet ontology \cite{fellbaum1998wordnet}.
However, all images have been resized to the ``tiny'' resolution of $32 \times 32$ pixels.

This low image resolution combined with the much more manageable size of the CIFAR datasets allows for fast training of CNNs.
Therefore, they have established themselves as one of the most popular benchmarks in the field of computer vision.

\section{Hunting Duplicates}
\label{sec:duplicate-search}

\subsection{Mining Duplicate Candidates}

\begin{figure}
    \includegraphics[width=\linewidth]{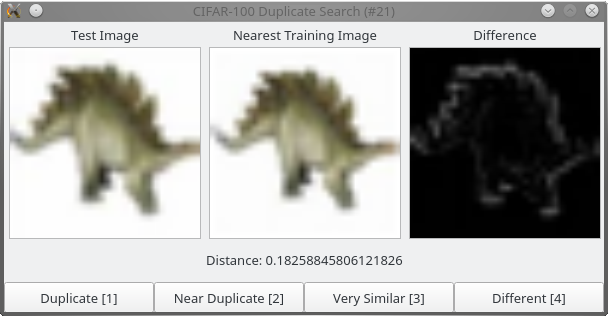}
    \caption{GUI for duplicate annotation.}
    \label{fig:gui-search}
\end{figure}

As we have argued above, simply searching for exact pixel-level duplicates is not sufficient, since there may also be slightly modified variants of the same scene that vary by contrast, hue, translation, stretching etc.
Thus, we follow a content-based image retrieval approach \cite{smeulders2000content,babenko2014neural,babenko2015aggregating} for finding duplicate and near-duplicate images:
We train a lightweight CNN architecture proposed by Barz et al. \cite{Barz18:GoodTraining} on the training set and then extract $L^2$-normalized features from the global average pooling layer of the trained network for both training and testing images.
Such classification-based features have been found to be useful for comparing images with regard to their visual similarity, and thus enjoy popularity for image retrieval applications \cite{babenko2014neural}.

For each test image, we find the nearest neighbor from the training set in terms of the Euclidean distance in that feature space.
Due to the $L^2$ normalization applied to the feature vectors, the Euclidean distance is equivalent to the cosine distance in this case, which is the distance metric used by most state-of-the-art image retrieval techniques \cite{babenko2015aggregating,husain2017rvd,revaud2019learning}.
Since we do not simply use features extracted from pre-trained models, but train the network on the training samples of the respective dataset, the feature representations are highly adapted to the domain covered by the images in the dataset.
Thanks to the data augmentation applied during training, the network sees many variants of each training image and is encouraged to learn features that are largely invariant against the kind of transformations that can result in near-duplicate images.

\subsection{Manual Annotation}

\begin{figure}[b!]
    \centering
    \includegraphics[width=\linewidth]{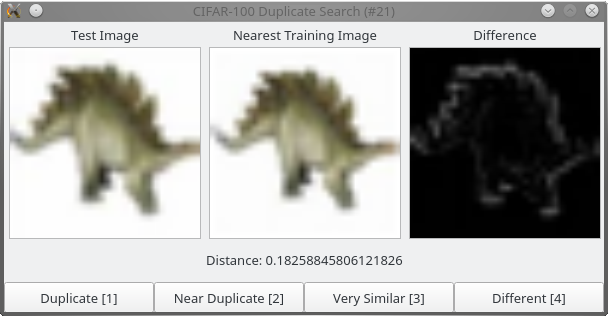}
    \caption{GUI for duplicate annotation.}
    \label{fig:gui-search}
\end{figure}

Given the nearest neighbors of each test image in the training set and their distance in the learned feature space, it would be easy to capture the majority of duplicates by simply thresholding the distance between these pairs.
However, such an approach would result in a high number of false positives as well.
Therefore, we inspect the detected pairs manually, sorted by increasing distance.
In a graphical user interface depicted in \cref{fig:gui-search}, the annotator can inspect the test image and its duplicate, their distance in the feature space, and a pixel-wise difference image.
The pair is then manually assigned to one of four classes:

\begin{description}
    \item[Exact Duplicate] Almost all pixels in the two images are approximately identical.
    \item[Near-Duplicate] The content of the images is exactly the same, \ie, both originated from the same camera shot. However, different post-processing might have been applied to this original scene, \eg, color shifts, translations, scaling etc.
    \item[Very Similar] The contents of the two images are different, but highly similar, so that the difference can only be spotted at the second glance.
    \item[Different] The pair does not belong to any other category.
\end{description}

\Cref{fig:dup-examples} shows some examples for the three categories of duplicates from the CIFAR-100 test set, where we picked the \nth{10}, \nth{50}, and \nth{90} percentile image pair for each category, according to their distance.
In the remainder of this paper, the word ``duplicate'' will usually refer to any type of duplicate, not necessarily to exact duplicates only.

\begin{figure}
    \includegraphics[width=\linewidth]{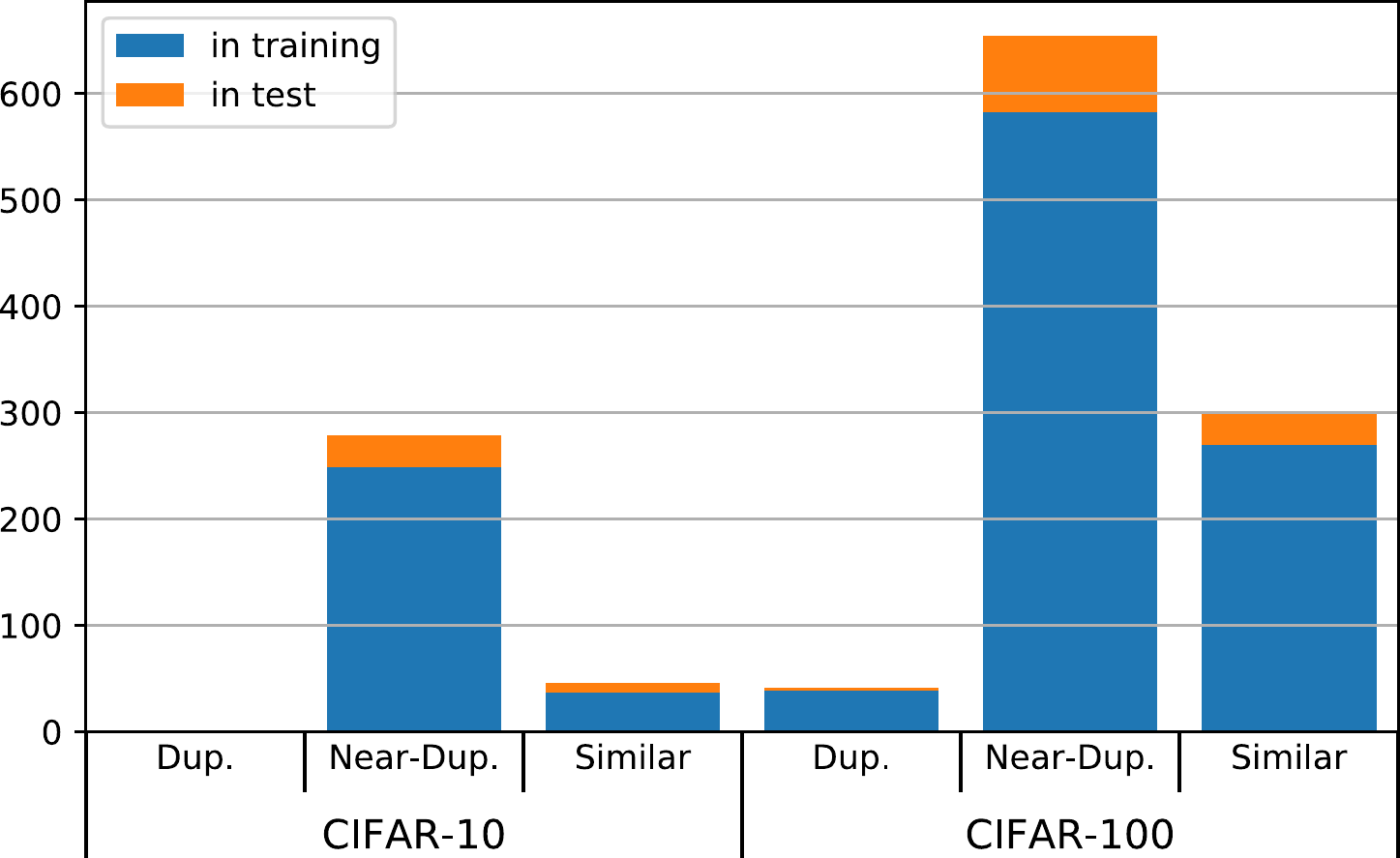}
    \caption{Number of duplicates per duplicate type between test and training set (blue) and within the test set (orange).}
    \label{fig:duplicate-type-distribution}
\end{figure}

We used a single annotator and stopped the annotation once the class ``Different'' has been assigned to 20 pairs in a row.
In addition to spotting duplicates of test images in the training set, we also search for duplicates within the test set, since these also distort the performance evaluation.
Note that we do not search for duplicates within the training set.
In the worst case, the presence of such duplicates biases the weights assigned to each sample during training, but they are not critical for evaluating and comparing models.

\subsection{Duplicate Statistics}

We found 891 duplicates from the CIFAR-100 test set in the training set and 104 duplicates within the test set itself.
In total, 10\% of the test images have duplicates.
The situation is slightly better for CIFAR-10, where we found 286 duplicates in the training and 39 in the test set, amounting to 3.25\% of the test set.
This is probably due to the much broader type of object classes in CIFAR-10:
We suppose it is easier to find 5,000 different images of birds than 500 different images of maple trees, for example.

The vast majority of duplicates belongs to the category of near-duplicates, as can be seen in \cref{fig:duplicate-type-distribution}.
It is worth noting that there are no exact duplicates in CIFAR-10 at all, as opposed to CIFAR-100.
This might indicate that the basic duplicate removal step mentioned by Krizhevsky et al.~\cite{krizhevsky2012alexnet} for the construction of CIFAR-10 has been omitted during the creation of CIFAR-100.

\Cref{tbl:top-dup-classes} lists the top 14 classes with the most duplicates for both datasets.
The only classes without any duplicates in CIFAR-100 are ``bowl'', ``bus'', and ``forest''.

\begin{table}
    \captionsetup[subtable]{position=top}
    \subcaptionbox{CIFAR-10}{%
        \renewcommand{\maxnum}{59}
        \renewcommand{\maxtext}{automobile}
        \renewcommand{\baralign}{l}
        \begin{tabular}{ l r }
            \databar{frog}{59} & 59 \\
            \databar{automobile}{55} & 55 \\
            \databar{airplane}{47} & 47 \\
            \databar{deer}{40} & 40 \\
            \databar{bird}{26} & 26 \\
            \databar{horse}{20} & 20 \\
            \databar{dog}{19} & 19 \\
            \databar{ship}{17} & 17 \\
            \databar{truck}{14} & 14 \\
            \databar{cat}{11} & 11 \\
            &\\&\\&\\&\\
        \end{tabular}%
    }\hfill%
    \subcaptionbox{CIFAR-100}{%
        \renewcommand{\maxnum}{33}
        \renewcommand{\maxtext}{lawn mower}
        \renewcommand{\baralign}{l}
        \begin{tabular}{ l r }
            \databar{cockroach}{33} & 33 \\
            \databar{orange}{29} & 29 \\
            \databar{lawn mower}{28} & 28 \\
            \databar{apple}{26} & 26 \\
            \databar{skunk}{26} & 26 \\
            \databar{keyboard}{23} & 23 \\
            \databar{leopard}{23} & 23 \\
            \databar{wolf}{21} & 21 \\
            \databar{worm}{20} & 20 \\
            \databar{whale}{17} & 17 \\
            \databar{rocket}{17} & 17 \\
            \databar{skyscraper}{17} & 17 \\
            \databar{dinosaur}{17} & 17 \\
            \databar{maple tree}{17} & 17 \\
        \end{tabular}%
    }%
    \caption{The top 14 classes with the most duplicates.}
    \label{tbl:top-dup-classes}
\end{table}

On the subset of test images with duplicates in the training set, the ResNet-110 \cite{he2016resnet} models from our experiments in \cref{sec:reevaluation} achieve error rates of 0\% and 2.9\% on CIFAR-10 and CIFAR-100, respectively.
These are in stark contrast to the 5.3\% and 26.1\% obtained by this model on the full test set, verifying that even near-duplicate and highly similar images can be classified far too easily by memorization.

\section{The Duplicate-Free ciFAIR Test Dataset}
\label{sec:new-test-set}

To create a fair test set for CIFAR, we replace all duplicates identified in the previous section with new images sampled from the Tiny Images dataset \cite{torralba2008tinyimages}, which was also the source for the original CIFAR datasets.

\begin{figure}
    \includegraphics[width=\linewidth]{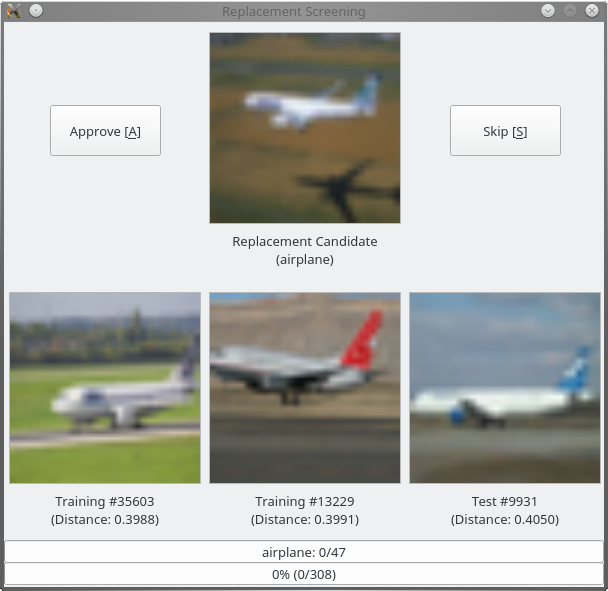}
    \caption{GUI for replacement candidate selection.}
    \label{fig:gui-replace}
\end{figure}

We took care not to introduce any bias or domain shift during the selection process.
To this end, each replacement candidate was inspected manually in a graphical user interface (see \cref{fig:gui-replace}), which displayed the candidate and the three nearest neighbors in the feature space from the existing training and test sets.
We approved only those samples for inclusion in the new test set that could not be considered duplicates (according to the definitions in \cref{sec:duplicate-search}) of any of the three nearest neighbors.

Furthermore, we followed the labeler instructions provided by Krizhevsky et al.~\cite{krizhevsky2009cifar} for CIFAR-10.
However, separate instructions for CIFAR-100, which was created later, have not been published.
By inspecting the data, we found that some of the original instructions seem to have been relaxed for this dataset.
For example, CIFAR-100 does include some line drawings and cartoons as well as images containing multiple instances of the same object category.
Both types of images were excluded from CIFAR-10.
Therefore, we also accepted some replacement candidates of these kinds for the new CIFAR-100 test set.

We term the datasets obtained by this modification as ciFAIR-10 and ciFAIR-100 (``fair CIFAR'').
They consist of the original CIFAR training sets and the modified test sets which are free of duplicates.
ciFAIR can be obtained online at \url{https://cvjena.github.io/cifair/}.

\section{Re-evaluation of the State of the Art}
\label{sec:reevaluation}

\begin{table*}
    \renewcommand{\arraystretch}{1.35}
    \rowcolors{2}{gray!10}{white}
    \begin{tabularx}{\linewidth}{X cccc cccc}
        \toprule
        \rowcolor{white}
        & \multicolumn{4}{c}{\textbf{CIFAR-10}} & \multicolumn{4}{c}{\textbf{CIFAR-100}} \\
        \cmidrule(r){2-5}\cmidrule(l){6-9}
        \textbf{Model} & \textbf{CIFAR} & \textbf{ciFAIR} & \textbf{Gap} & \textbf{Rel.~Gap} & \textbf{CIFAR} & \textbf{ciFAIR} & \textbf{Gap} & \textbf{Rel.~Gap} \\
        \specialrule{1pt}{.4ex}{0pt}
        Plain-11 \cite{Barz18:GoodTraining} & 5.91\% & 6.43\% & 0.52\% & 8.80\% & 27.82\% & 31.34\% & 3.52\% & 12.65\% \\
        ResNet-110 \cite{he2016resnet} & 5.26\% & 5.77\% & 0.51\% & 9.70\% & 26.05\% & 29.24\% & 3.19\% & \textbf{12.25\%} \\
        WRN-28-10 \cite{zagoruyko2016wrn} & 3.89\% & 4.25\% & 0.36\% & 9.25\% & 18.95\% & 21.48\% & 2.53\% & 13.35\% \\
        DenseNet-BC (L=190, k=40) \cite{huang2017densenet} & 3.90\% & 4.20\% & \textbf{0.30\%} & \textbf{7.69\%} & 18.62\% & 21.02\% & 2.40\% & 12.89\% \\
        ResNeXt-29 (8x64d) \cite{xie2017resnext} & \textbf{3.56\%} & \textbf{3.95\%} & 0.39\% & 10.96\% & 18.38\% & 20.84\% & 2.46\% & 13.38\% \\
        PyramidNet-272-200 \cite{han2017pyramidnet} & 3.58\% & 4.00\% & 0.42\% & 11.73\% & \textbf{17.05\%} & \textbf{19.38\%} & \textbf{2.33\%} & 13.67\% \\
        \specialrule{1pt}{0pt}{0pt}
    \end{tabularx}
    \caption{Classification error rate of various CNN architectures on the original CIFAR test sets and the modified ciFAIR test sets. The best value in each column is highlighted in bold font.}
    \label{tbl:performance}
\end{table*}

Two questions remain:
Were recent improvements to the state-of-the-art in image classification on CIFAR actually due to the effect of duplicates, which can be memorized better by models with higher capacity?
Does the ranking of methods change given a duplicate-free test set?

To answer these questions, we re-evaluate the performance of several popular CNN architectures on both the CIFAR and ciFAIR test sets.
Unfortunately, we were not able to find any pre-trained CIFAR models for any of the architectures.
Thus, we had to train them ourselves, so that the results do not exactly match those reported in the original papers.
However, we used the original source code, where it has been provided by the authors, and followed their instructions for training (\ie, learning rate schedules, optimizer, regularization etc.).

The results are given in \cref{tbl:performance}.
Besides the absolute error rate on both test sets, we also report their difference (``gap'') in terms of absolute percent points, on the one hand, and relative to the original performance, on the other hand.

On average, the error rate increases by 0.41 percent points on CIFAR-10 and by 2.73 percent points on CIFAR-100.
The relative difference, however, can be up to 12\%.
The ranking of architectures is stable on CIFAR-100, and only WRN and DenseNet swapped positions on CIFAR-10.

\section{Conclusions}
\label{sec:conclusions}

In a laborious manual annotation process supported by image retrieval, we have identified a startling number of duplicate images in the CIFAR test sets that also exist in the training set.
We have argued that it is not sufficient to focus on exact pixel-level duplicates only.
In contrast, slightly modified variants of the same scene or very similar images bias the evaluation as well, since these can easily be matched by CNNs using data augmentation, but will rarely appear in real-world applications.
We hence proposed and released a new test set called ciFAIR, where we replaced all those duplicates with new images from the same domain.

A re-evaluation of several state-of-the-art CNN models for image classification on this new test set led to a significant drop in performance, as expected.
In fact, even simple models such as ResNet-110 classified almost all the near-duplicate images perfectly, achieving error rates on this subset that are an order of magnitude better than on the remaining non-duplicate images.
This observation underpins that the possibility of memorizing the training data is heavily exploited by state-of-the-art CNN architectures.
In combination with the high fraction of near-duplicates in the test set (as high as 10\% on CIFAR-100), this effect has the potential to bias the evaluation and development of deep learning approaches for image recognition substantially.

A positive result, on the other hand, is that the relative ranking of the models did not change considerably.
Thus, the research efforts of the community do not seem to have overfitted to the presence of duplicates in the test set yet:
CNN architectures performing better than others on the original test still provide higher accuracy on the new ciFAIR test set.
However, all models we tested have sufficient capacity to memorize the complete training data.
In the past, the high fraction of near-duplicates in the test data might hence have prevented the publication of less over-parameterized approaches that are computationally more efficient and might generalize just as well as other models, but fail to memorize the duplicates.

We encourage all researchers training models on the CIFAR datasets to evaluate their models on ciFAIR, which will provide a better estimate of how well the model generalizes to new data.
To facilitate comparison with the state-of-the-art further, we maintain a community-driven leaderboard at \url{https://cvjena.github.io/cifair/}, where everyone is welcome to submit new models.
We will only accept leaderboard entries for which pre-trained models have been provided, so that we can verify their performance.

{\small
\bibliographystyle{ieee}
\bibliography{references}

\begin{thebibliography}{10}\itemsep=-1pt

\bibitem{babenko2015aggregating}
A.~Babenko and V.~Lempitsky.
\newblock Aggregating local deep features for image retrieval.
\newblock In {\em {IEEE} International Conference on Computer Vision (ICCV)},
  pages 1269--1277, 2015.

\bibitem{babenko2014neural}
A.~Babenko, A.~Slesarev, A.~Chigorin, and V.~Lempitsky.
\newblock Neural codes for image retrieval.
\newblock In {\em European Conference on Computer Vision (ECCV)}, pages
  584--599. Springer, 2014.

\bibitem{Barz18:GoodTraining}
B.~Barz and J.~Denzler.
\newblock Deep learning is not a matter of depth but of good training.
\newblock In {\em International Conference on Pattern Recognition and
  Artificial Intelligence (ICPRAI)}, pages 683--687. CENPARMI, Concordia
  University, Montreal, 2018.

\bibitem{deng2009imagenet}
J.~Deng, W.~Dong, R.~Socher, L.-J. Li, K.~Li, and L.~Fei-Fei.
\newblock {I}mage{N}et: A large-scale hierarchical image database.
\newblock In {\em IEEE Conference on Computer Vision and Pattern Recognition
  (CVPR)}, pages 248--255. {IEEE}, 2009.

\bibitem{fellbaum1998wordnet}
C.~Fellbaum.
\newblock {\em WordNet}.
\newblock Wiley Online Library, 1998.

\bibitem{han2017pyramidnet}
D.~Han, J.~Kim, and J.~Kim.
\newblock Deep pyramidal residual networks.
\newblock In {\em IEEE Conference on Computer Vision and Pattern Recognition
  (CVPR)}, pages 6307--6315. IEEE, 2017.

\bibitem{he2016resnet}
K.~He, X.~Zhang, S.~Ren, and J.~Sun.
\newblock Deep residual learning for image recognition.
\newblock In {\em IEEE Conference on Computer Vision and Pattern Recognition
  (CVPR)}, pages 770--778, 2016.

\bibitem{huang2017densenet}
G.~Huang, Z.~Liu, L.~Van Der~Maaten, and K.~Q. Weinberger.
\newblock Densely connected convolutional networks.
\newblock In {\em IEEE Conference on Computer Vision and Pattern Recognition
  (CVPR)}, pages 4700--4708, 2017.

\bibitem{huiskes08mirflickr}
M.~J. Huiskes and M.~S. Lew.
\newblock The {MIR Flickr} retrieval evaluation.
\newblock In {\em MIR '08: Proceedings of the 2008 ACM International Conference
  on Multimedia Information Retrieval}, New York, NY, USA, 2008. ACM.

\bibitem{husain2017rvd}
S.~S. Husain and M.~Bober.
\newblock Improving large-scale image retrieval through robust aggregation of
  local descriptors.
\newblock {\em IEEE Transactions on Pattern Analysis and Machine Intelligence},
  39(9):1783--1796, 2017.

\bibitem{jaderberg2015spatial}
M.~Jaderberg, K.~Simonyan, A.~Zisserman, and K.~Kavukcuoglu.
\newblock Spatial transformer networks.
\newblock In C.~Cortes, N.~D. Lawrence, D.~D. Lee, M.~Sugiyama, and R.~Garnett,
  editors, {\em Advances in Neural Information Processing Systems (NIPS)},
  pages 2017--2025. Curran Associates, Inc., 2015.

\bibitem{krizhevsky2009cifar}
A.~Krizhevsky and G.~Hinton.
\newblock Learning multiple layers of features from tiny images.
\newblock Technical report, University of Toronto, 2009.

\bibitem{krizhevsky2012alexnet}
A.~Krizhevsky, I.~Sutskever, and G.~E. Hinton.
\newblock {I}mage{N}et classification with deep convolutional neural networks.
\newblock In {\em Advances in Neural Information Processing Systems (NIPS)},
  pages 1097--1105, 2012.

\bibitem{real2018amoebanet}
E.~Real, A.~Aggarwal, Y.~Huang, and Q.~V. Le.
\newblock Regularized evolution for image classifier architecture search.
\newblock In {\em Proceedings of the aaai conference on artificial
  intelligence}, volume~33, pages 4780--4789, 2019.

\bibitem{recht2018cifar}
B.~Recht, R.~Roelofs, L.~Schmidt, and V.~Shankar.
\newblock Do cifar-10 classifiers generalize to cifar-10?
\newblock {\em arXiv preprint arXiv:1806.00451}, 2018.

\bibitem{revaud2019learning}
J.~Revaud, J.~Almazan, R.~S. Rezende, and C.~R.~d. Souza.
\newblock Learning with average precision: Training image retrieval with a
  listwise loss.
\newblock In {\em {IEEE} International Conference on Computer Vision (ICCV)},
  pages 5107--5116, October 2019.

\bibitem{russakovsky2015ilsvrc}
O.~Russakovsky, J.~Deng, H.~Su, J.~Krause, S.~Satheesh, S.~Ma, Z.~Huang,
  A.~Karpathy, A.~Khosla, M.~Bernstein, et~al.
\newblock {I}mage{N}et large scale visual recognition challenge.
\newblock {\em International Journal of Computer Vision}, 115(3):211--252,
  2015.

\bibitem{smeulders2000content}
A.~W. Smeulders, M.~Worring, S.~Santini, A.~Gupta, and R.~Jain.
\newblock Content-based image retrieval at the end of the early years.
\newblock {\em IEEE Transactions on Pattern Analysis and Machine Intelligence
  (TPAMI)}, 22(12):1349--1380, 2000.

\bibitem{sun2017revisiting}
C.~Sun, A.~Shrivastava, S.~Singh, and A.~Gupta.
\newblock Revisiting unreasonable effectiveness of data in deep learning era.
\newblock In {\em IEEE International Conference on Computer Vision (ICCV)},
  pages 843--852. IEEE, 2017.

\bibitem{torralba2008tinyimages}
A.~Torralba, R.~Fergus, and W.~T. Freeman.
\newblock 80 million tiny images: A large data set for nonparametric object and
  scene recognition.
\newblock {\em IEEE Transactions on Pattern Analysis and Machine Intelligence
  (TPAMI)}, 30(11):1958--1970, 2008.

\bibitem{WahCUB_200_2011}
C.~Wah, S.~Branson, P.~Welinder, P.~Perona, and S.~Belongie.
\newblock {The Caltech-UCSD Birds-200-2011 Dataset}.
\newblock Technical Report CNS-TR-2011-001, California Institute of Technology,
  2011.

\bibitem{tencent-ml-images-2019}
B.~Wu, W.~Chen, Y.~Fan, Y.~Zhang, J.~Hou, J.~Huang, W.~Liu, and T.~Zhang.
\newblock {T}encent {ML}-{I}mages: A large-scale multi-label image database for
  visual representation learning.
\newblock {\em {IEEE} Access}, 7:172683--172693, 2019.

\bibitem{xie2017resnext}
S.~Xie, R.~Girshick, P.~Doll{\'a}r, Z.~Tu, and K.~He.
\newblock Aggregated residual transformations for deep neural networks.
\newblock In {\em IEEE Conference on Computer Vision and Pattern Recognition
  (CVPR)}, pages 5987--5995. IEEE, 2017.

\bibitem{zagoruyko2016wrn}
S.~Zagoruyko and N.~Komodakis.
\newblock Wide residual networks.
\newblock In E.~R.~H. Richard C.~Wilson and W.~A.~P. Smith, editors, {\em
  British Machine Vision Conference (BMVC)}, pages 87.1--87.12. BMVA Press,
  September 2016.

\end{thebibliography}
}

\end{document}